\documentclass{llncs}
\usepackage{graphicx}
\usepackage{multicol}
\usepackage{amsmath}
\usepackage{makeidx}  
\newtheorem{Def}{Definition}

\begin{document}
%
%
\title{Inheritance in Object-Oriented Knowledge Representation}
\titlerunning{Inheritance in Object-Oriented Knowledge Representation}  
%
\author{Dmytro Terletskyi}
\authorrunning{} 
%
\tocauthor{}
\institute{Taras Shevchenko national University of Kyiv, Kyiv, 03680, Ukraine
\email{dmytro.terletskyi@gmail.com},\\
\texttt{http://cyb.univ.kiev.ua/en/departments.is.terletskyi.html}
}

\maketitle              

\begin{abstract}
This paper contains the consideration of inheritance mechanism in such knowledge representation models as object-oriented programming, frames and object-oriented dynamic networks. In addition, inheritance within representation of vague and imprecise knowledge are also discussed. New types of inheritance, general classification of all known inheritance types and approach, which allows avoiding in many cases problems with exceptions, redundancy and ambiguity within object-oriented dynamic networks and their fuzzy extension, are introduced in the paper. The proposed approach bases on conception of homogeneous and inhomogeneous or heterogeneous class of objects, which allow building of inheritance hierarchy more flexibly and efficiently. 
\end{abstract}
\section{Introduction}
Nowadays the design and development of knowledge-based systems for solving problems in different domains are important tasks within area of artificial intelligence. Currently there are many different knowledge representation models (KRM), the most famous of which are logical models, production models, semantic networks, frames, scripts, conceptual graphs, ontologies, etc. All of these KRMs have their own specifics and allow representing of some types of knowledge. However, the certain programming paradigm should be chosen for implementation of any particular KRM. For today the most famous and commonly used programming paradigm is an object-oriented programming (OOP). It gives us an opportunity of efficient implementation of many existing KRM, in particular those that are object-oriented, e.g. frames, scripts. We should take into account that the knowledge in forms of any KRM must be somehow represented in the database. Object-oriented approach to knowledge representation is very suitable for this purpose, because it provides such powerful tool, as inheritance mechanism. It allows building of inheritance hierarchies and avoiding of redundancy of knowledge representation in database, because it partially implements the conception of reusability. In its turn, inheritance hierarchy as a type of knowledge structure provides an efficient mechanisms of reasoning about knowledge. Furthermore, the modern versions of most OOP-languages support such a programming technique as object-relational mapping (ORM), which provides convenient interaction among object-oriented programs and databases.

However, despite all advantages of object-oriented approach to knowledge representation, it also has some drawbacks. Firstly, inheritance mechanism leads three main kinds of problems, such as \emph{problem of exceptions}, \emph{problem of redundancy} and \emph{problem of ambiguity} \cite{Al-Asady}, \cite{Touretzky}. They arise during constructing of inheritance hierarchies and reasoning within them. Secondly, a lot of human knowledge have vague and imprecise nature \cite{Berzal}, \cite{Leung-Wong} and OOP does not support representation of such knowledge. Thirdly, OOP provides an opportunity to create and to operate only with homogeneous classes \cite{Terletskyi-1}, that is why we need to create new class for every new type of objects, even when some of them are similar.

\section{Inheritance in Object-Oriented Programming}
Nowadays there are two main approaches in modern OOP, which are implemented within \emph{class-based} and \emph{prototype-based} programming languages \cite{Craig}. The main idea of first approach is an identification of common properties of some quantity of objects and their description within such structure as class. Objects exist only in runtime as a result of instantiating of a class. Within the second approach, the objects are results of cloning operation, which is applied to \emph{prototypes}, where prototypes define stereotypical objects. The new prototype can be obtained as a modification of copy of other prototype. Currently, class-based programming approach is more commonly used than prototype-based one and most of modern OOP-languages support exactly class-based style. That is why all future considerations concerning OOP will be done within class-based programming approach.

In paradigm of OOP, class defines a kind of a concept, and objects are instances of it. Each class consists of fields and methods, where fields define the structure of the class and methods define its behavior. In other words, fields define properties of the concept and methods are functions that give an opportunity to manipulate them. When the program creates an object as an instance of some class, this object has the same fields, as its class and each method of the class can be called for this object. In such a way, class implements the mechanism of encapsulation, because the object has the same structure and behavior, but it has its own values of the fields, which can differ from corresponding values of class's fields and can be changed during program execution.

\subsection{Single Inheritance}
Class-based approach provides an ability to define the class using the existing definition of another class. In this case, one class can inherit specifics of another one. Moreover, it can extend or specialize the inherited specifics by adding its own features. This process is called \emph{single inheritance} \cite{Craig}. Using this mechanism, we can build inheritance hierarchies, where concepts that are more general will have higher position in the hierarchy than those that are less general. Class which inherits another class is called a \emph{subclass} of that class and the class, which was inherited by another class, is called a \emph{superclass}. Single inheritance can be graphically represented as a tree. 

According to \cite{Craig} there are at least three different interpretations of inheritance. We will consider inheritance in the context of modeling of classification hierarchies in the chosen application domain. Such interpretation is more common in OOP and is used in object-oriented knowledge representation.

Proposed approach has some benefits. Usage of inheritance allows more efficient using of computer memory and memory in a database by avoiding duplication of similar information, during description of classes. Almost all modern OOP-languages support single inheritance. However, it also has some drawbacks. When one class inherits another one, it inherits all its properties. There are some cases when it causes some redundancy of description of subclasses, moreover sometimes it causes conflicts among concepts, described by subclass and superclass. All these problems will be considered and discussed in more detail later.

\subsection{Multiple Inheritance}
Under single inheritance, each subclass can have only one superclass, however class can have more than one superclass and there are cases when single inheritance is insufficient. For this purpose there is another form of inheritance, which is called an \emph{multiple inheritance} \cite{Craig}. It allows class to inherit specifics of many other classes. Multiple inheritance hierarchy can be graphically represented as an acyclic directed graph, or simply an direct inheritance graph.

Multiple inheritance has almost the same benefits, as a single one. Moreover, it gives an opportunity to create more complex classes and objects via inheritance. However, multiple inheritance also has some drawbacks. Usage of multiple inheritance sometimes causes two types of semantic conflicts within the subclasses. In the first case, the class can simultaneously inherit a few copies of the same method or different values of the equivalent properties from different superclasses. In the second case, the subclass can inherit semantically incompatible properties and methods. In addition, in contrast to single inheritance, not every OOP-language supports multiple inheritance. Languages, which support multiple inheritance, are C++, Common Lisp, Eiffel, Scala, Perl, Python, etc. However, for example such commonly used OOP-languages as C\#, Java, Objective-C, Ruby, Php do not do it. Most of them use an alternative approaches to multiple inheritance such as interfaces, that allow partial 
modelling of multiple inheritance principles.

\section{Object-Oriented Knowledge Representation}
The main idea of object-oriented knowledge representation approach is representation of knowledge about a domain in terms of objects, classes and relations among them. OOP provides all opportunities for such representation, however in many books where models of knowledge representation are described, OOP is not mentioned. Nevertheless, we consider models, which are ideologically close to OOP, such as frames and object-oriented dynamic networks (OODN). We briefly consider these KRMs and implementation of inheritance mechanism within them. 

\subsection{Frames}
Frame is a data-structure for representation of knowledge about stereotypical situations \cite{Minsky}. Frame consists of set of slots, where each slot has its own filler. Name of a frame, relationships with other frames, attributes of frame, procedural attachments can be fillers for frame's slots. Every slot with its value represents particular property of object or class, which is represented by frame. Generally, there are two types of frames: individual or \emph{instance-frames} for representation of single objects, and generic or \emph{class-frames} for representation of classes \cite{Negnevitsky}. 

Different frames can be merged into one system via relationships \cite{Minsky}. There are three main types of relations among frames: \emph{generalization}, \emph{aggregation} and \emph{association}. Generalization represents relationship between subclass and superclass or object and class, when subclass is a kind of superclass or object is an instance of its class. This type of relationships can be denoted using \emph{is-a}, \emph{an-instance-of}, \emph{a-kind-of}, etc. links. Aggregation represents relationship among several subclasses and their superclass, when subclass is a part of superclass. Usually aggregation can be denoted as \emph{a-part-of}, \emph{part-whole}, etc. Association describes some semantic relationship among different classes, which are unrelated otherwise. Examples of such kind relationships are \emph{have}, \emph{can}, \emph{own}, etc.

Usually, frames can have some methods associated with them. They are called procedural attachments. Every procedure is a set of some instructions, which are associated with a frame and can be executed on request. 

Similarly to OOP, frames use the inheritance mechanism for building frames-systems, which also have hierarchical structure \cite{Negnevitsky}. The conception of inheritance within frames is the same as in OOP. There is difference only between representation of structure of classes and objects within these approaches. OOP is more flexible and powerful for representation of class structure, because in contrast to frames, it has some set of basic built-in primitive data types, which can be used for creating more complex data structures, while frames has only three built-in primitive types: numeric, string and logical. However, frames have such feature as compound attributes which take a value from some set of values, which elements can have different types. 

In terms of frames, class can inherit specifics of another class through generalization relationship, i.e. \emph{is-a} slot. However, single and multiple inheritance cause the same problems in frames as in OOP \cite{Negnevitsky}, \cite{Touretzky}.

As we can see, problems of inheritance are common for all object-oriented KRMs, but they are related only to the specifics of inheritance mechanism.

\subsection{Problem of Exceptions}
The first known problem of inheritance is the problem of exceptions. There are some classical examples, which illustrate it. They are known as examples about flying penguins or ostriches and about three-legged or white elephant \cite{Al-Asady}, \cite{Touretzky}. In general, the problem can be formulated as a situation, when superclass contains properties, which are not true for all its subclasses. 

After formulation of this problem, a few approaches to its solving were proposed. For example, in frame-based systems, subclasses can override the values of inherited slots from their superclass \cite{Negnevitsky}. However, this approach is not efficient, because overriding of values of slots leads to the situation when the subclass goes beyond its superclass. After it, this class cannot be viewed as the subclass of its superclass, because all subclasses must inherit all properties of their superclass. 

The main idea of another known approach is the usage of \emph{not-is-a} links for modelling of exceptions \cite{Al-Asady}, \cite{Touretzky}. Such solution differs from others, because its main idea is not to avoid the exceptions in the hierarchy, but to describe them somehow. The conception of \emph{not-is-a} link came from logical approach of knowledge representation and on the first glance such solution does not cause any suspicions. However it causes appearing of the contradictory classes, formation of inconsistent knowledge base and as result contradictory reasoning \cite{Al-Asady}.

In OOP, solving of this problem relies on the programmer. In other words, the programmer should somehow constrain the generality of the superclass. 

\subsection{Problem of Redundancy}
One more kind of problem related to inheritance is the problem of redundancy. It appears within the inheritance tree, when the class inherits specifics from more than one related superclass \cite{Al-Asady}, \cite{Touretzky}. In this situation, there is a vertical chain of inheritance, where top level contains most general class and each of lower levels contains less general class, than its superclass. On the bottom level there is the most specific class of the hierarchy. The main features of this class is that it inherits all properties from its predecessors. Sometimes such inheritance is redundant, because the class can inherit unnecessary properties or methods and all objects of this class will have the same specifics.

There are some approaches, which avoid the inheritance of redundant properties. One of them is the choosing of the nearest value. However, it is not an efficient way, because the result of such choosing depends on appropriate algorithm. Various systems have different algorithms, which can return different results in the same situation \cite{Al-Asady}, \cite{Touretzky}.

\subsection{Problem of Ambiguity}
Another kind of problem related to inheritance is the problem of ambiguity. There are a few classical examples, which illustrate this problem. They are known as examples about Quaker or Nixon and about elephant, who is a circus performer, etc. \cite{Al-Asady}, \cite{Touretzky}. This problem appears, when the class inherits specifics from more than one unrelated superclass of the same level, and these superclasses contain properties and methods with the same names. In this situation, subclass should somehow choose one of these variants. 

Concerning properties, sometimes they can have only similar name, but not a type or value. Sometimes, they can have the same type and different values or they can have the same type and value. In all these cases there is an ambiguity, because it is unknown, which particular property should be chosen and different variants can have totally different semantic contexts. Methods, similarly to properties can have only the same names and very different semantic contexts. However, even if their semantics are similar or close to similar, they can be implemented in different ways.

There are a few approaches for solving this problem \cite{Al-Asady}, \cite{Touretzky}. First of them uses the idea of choosing some particular version of property or method. In this case, there is a question how to choose them. There are appropriate algorithms, which are implemented in different systems, in particular in frame-based ones. However, they use different criteria for choosing the variant. Very often result depends on the behavior and time complexity of the algorithm. It means that different algorithms will give different results using the same inheritance structure. Second approach allows inheritance of all possible variations of properties and methods. In this situation results will be different in various systems \cite{Al-Asady}. However, both solutions are not efficient enough, because in the first case a system ignores some part of variants in different ways and in the second one, knowledge base becomes inconsistent.

\subsection{Object-Oriented Dynamic Networks}
Another kind of object-oriented knowledge representation model is object-oriented dynamic networks, which was proposed in \cite{Terletskyi-2}. In some aspects, this KRM is similar to OOP and frames, however, despite this, it has some specific peculiarities, which are not typical for other models. Let us consider structure of this model.
\begin{Def}
Object-Oriented Dynamic Network is a 5-tuple 
\[OODN=(O,C,R,E,M),\] 
where:
\begin{itemize}
 \item $O$ -- a set of objects;
 \item $C$ -- a set of classes of objects, which describe objects from set $O$;
 \item $R$ -- a set of relations, which are defined on set $O$ and $C$;
 \item $E$ -- a set of exploiters, which are defined on set $O$ and $C$;
 \item $M$ -- a set of modifiers, which are defined on set $O$ and $C$.
\end{itemize}
\end{Def}
Analyzing this definition, we can conclude that usage of conceptions of objects, classes and relation among them is common for both OOP and frames. However, all these concepts have different implementations within mentioned KRMs. One of the main differences is the definition of the class. Within OOP, class is something like abstract description of some quantity of objects of the same nature \cite{Craig}. According to this, such class is homogeneous, because all its instances have the same type. In this sense, definition of the class within frames is similar to appropriate one in OOP. However, there is another type of classes, which are inhomogeneous or heterogeneous \cite{Terletskyi-1}. Conception of a class, which is defined within OODN, takes into account both types of classes. Let us consider it in more details.
\begin{Def}
\label{def-2}
Class of objects $T$ is a tuple $T=(P(T),F(T))$, where $P(T)$ is specification (a set of properties) of some quantity of objects, and $F(T)$ is their signature (a set of methods).
\end{Def}
The next definition proposes some classification of classes.
\begin{Def}
Homogeneous class of objects is a class of objects, which contains only similar objects.
\end{Def}
According to this, we can conclude that definition \ref{def-2} describes homogeneous classes.

Now let us consider the definition of inhomogeneous class.
\begin{Def}
Inhomogeneous (heterogeneous) class of objects $T$ is a tuple 
\[T=(Core(T),pr_1(A_1),\dots,pr_n(A_n)),\] 
where $Core(T)=(P(T),F(T))$ is the core of class of objects $T$, which includes only properties and methods similar to corresponding properties of specifications $P(A_1),\dots,P(A_n)$ and corresponding methods of signatures $F(A_1),\dots,F(A_n)$ respectively, and where $pr_i(A_i)=(P(A_i),F(A_i))$ , $i=\overline{1,n}$ are projections of objects $A_1,\dots,A_n$, which consist of properties and methods typical only for these objects.
\end{Def}
This approach gives an opportunity to describe some quantity of objects, which have similar or even different nature within one class. While in OOP, we must define new class for each new type of objects, even if these types are close or similar. 

Some of main features of OODN are a set of exploiters $E$ and a set of modifiers $M$. Both of them contain methods which can be applied to the objects and classes of objects from set $O$ and $C$ respectively. The difference between these two types of methods is character of their action. Exploiters use the objects and classes of objects, as the parameters for obtaining new knowledge, without any their changes, while, modifiers change the essence of objects and classes of objects and allow modelling of changes of basic knowledge over time.

In general, OODN can be viewed as two conceptual parts. First of them is declarative, which includes sets $O$, $C$, $R$, and allows representation of knowledge about particular domain. Second part is procedural one, it includes sets $E$, $M$ and provides the tools for obtaining new knowledge from basic ones.

\section{Object-Oriented Representation of Fuzzy Knowledge}
Currently there is variety of KRMs, which give an opportunity to represent the knowledge in different ways. Main of them were mentioned in the introduction part. However, a lot of human knowledge is vague and imprecise \cite{Berzal}, \cite{Leung-Wong} and cannot be represented in efficient way, using existing KRMs. That is why most of them were extended to the case of fuzzy knowledge, through the use of fuzzy sets theory \cite{Zadeh}. Currently there are fuzzy logic, fuzzy semantic networks, fuzzy rule-based models, fuzzy neural networks, fuzzy ontologies, fuzzy frames, fuzzy UML, etc. However, classical paradigm of OOP does not provide an opportunity for representing fuzzy objects and classes. That is why, a few attempts to do this were done within object-oriented approach to representation of fuzzy knowledge \cite{Berzal}, \cite{Leung-Wong}, \cite{Ndousse}. 

Similarly to object-oriented knowledge representation, main concepts of object-oriented representation of fuzzy knowledge are fuzzy objects, classes of fuzzy objects and relationships among them. The object and class are fuzzy, when they have at least one fuzzy property, i.e. property that is defined by a fuzzy set. The relations among fuzzy objects and classes of fuzzy objects, which are usually considered are similar to corresponded relations in frames and OOP, i.e. generalization, aggregation and association.

\subsection{Fuzzy Frames}
One of the most interesting extensions of classical KRMs to the case of fuzzy knowledge are fuzzy frames \cite{Graham-Jones-1}, \cite{Graham-Jones-2}. There are two main differences between frames and fuzzy frames. Firstly, within fuzzy frames slots can contain fuzzy sets as values. Secondly, the inheritance through \emph{is-a} slot can be partial. Such extension of frames allows describing of objects and classes which have partial properties, i.e. properties which inherent with some measure. It means that such properties are not strictly true or false for the object or class. This kind of inheritance is called weaker inheritance.

Proposed kind of inheritance can solve problem with exceptions in some cases, when the subclass inherits all properties of its superclass, but some of them are inherited with measure less than 1. It means that these properties are less expressed within the subclass than in its ancestor. However, such approach does not solve problems with redundancy and ambiguity, because it allows only the flexible description of inheritance relationships among classes.

\subsection{Fuzzy Object-Oriented Dynamic Networks}
Similarly to OOP and frames, object-oriented-dynamic networks are not efficient for representation of fuzzy objects and classes. That is why concepts of fuzzy object, class of fuzzy objects which are basic for OODN were extended to the case of fuzzy knowledge \cite{Terletskyi-3}. Taking into account these extensions, the definition of fuzzy object-oriented dynamic networks can be formulated in the following way.
\begin{Def}
Fuzzy Object-Oriented Dynamic Network is a object-oriented dynamic network
\[FOODN=(O,C,R,E,M),\] 
for which at least one of the following conditions:
\begin{itemize}
 \item $\exists A_k,\dots,A_m\in O=\{A_1,\dots,A_n\}$, where $1\leq k\leq m\leq n$ and $A_k,\dots,A_m$ are fuzzy objects;
 \item $\exists T_p,\dots,T_q\in C=\{T_1,\dots,T_w\}$, where $1\leq p\leq q\leq w$ and $T_p,\dots,T_q$ are classes of fuzzy objects;
 \item $\exists R_i,\dots,R_j\in R=\{R_1,\dots,R_v\}$, where $1\leq i\leq j\leq v$ and $R_i,\dots,R_j$ are fuzzy relations among fuzzy objects and classes of fuzzy objects.
\end{itemize}
is true.
\end{Def}
The most important feature of this extension is that general structure of the object and class of objects, types of classes and relations are the same for OODN and FOODN. There only difference is the type of properties, because in FOODN some properties of objects or classes of objects can be fuzzy.

\section{Types of Inheritance}
As we can see from previous sections, there are two types of inheritance -- single and multiple. Such inheritance types classification allows consideration of inheritance process in the context of different types of inheritance source. However, there is another classification, which divides inheritance on strong and weak. It allows consideration of inheritance from another point of view, namely how the inherited properties will be expressed within the subclass.

Nevertheless, there are other classifications. The common feature for single and multiple inheritance is that subclass inherits all properties and methods of inheritance source. We suppose that it is the source of majority of problems. In our opinion, if the class did not inherit all the properties of inheritance source, it would not cause the problems of redundancy and ambiguity. Moreover, such kind of inheritance allows building of inheritance hierarchy in more flexible way, without redundancy and ambiguity. According to this, we can conclude that inheritance can be also classified as full and partial. In the first case subclass inherits all the properties and methods from inheritance source, in the second case it inherits only selected properties and methods. All considered classifications of inheritance can be arranged within one classification, which is represented in the Table~\ref{tab-1}.
\begin{table}
\centering{
\caption{Classification of Inheritance Types}
\label{tab-1}
\begin{tabular}{cccccccc}
\hline\noalign{\smallskip}
\multicolumn{8}{c}{Inheritance}\\
\noalign{\smallskip}
\hline
\noalign{\smallskip}
\multicolumn{4}{c}{Single} & \multicolumn{4}{c}{Multiple}\\
\multicolumn{2}{c}{Full} & \multicolumn{2}{c}{Partial} & \multicolumn{2}{c}{Full} & \multicolumn{2}{c}{Partial}\\
Strong\ \ & Weak\ \ & Strong\ \ & Weak\ \ & Strong\ \ & Weak\ \ & Strong\ \ & Weak\ \ \\
\hline
\end{tabular}}
\end{table}

Now, let us consider the process of inheritance within OODN and FOODN. Suppose we have three classes of object $A_1$, $A_2$ and $A_3$, which are defined as follows
\begin{align*}
& T(A_1)=(P(A_1),F(A_1))=(p_1(A_1),p_2(A_1),f_1(A_1),f_2(A_1)),\\
& T(A_2)=(P(A_2),F(A_2))=(p_1(A_2),p_2(A_2),f_1(A_2)),\\
& T(A_3)=(P(A_3),F(A_3))=(p_1(A_3),f_1(A_3)).
\end{align*}
Let us consider types of inheritance, which are shown in Table~\ref{tab-1}, within OODN and FOODN. According to Table~\ref{tab-1}, there are eight different types of inheritance, but all of them can be reduced to two main kinds -- single and multiple. That is why, let us consider these two types as the most general ones.

\subsection{Single Inheritance}
Suppose we have the following sequence of inheritance
\[A_3\xrightarrow[]{inherits}A_2\xrightarrow[]{inherits}A_1.\]
The result of such inheritance is 
\[A_3\xrightarrow[]{inherits}A_2\xrightarrow[]{inherits}A_1=T=(Core(T),pr_1(T),pr_1(pr_1(T))),\]
where
\begin{align*}
Core(T)&=(p_1(A_1),p_2(A_1),f_1(A_1),f_2(A_1)),\\
pr_1(T)&=(p_1(A_2),p_2(A_2),f_1(A_2)),\\
pr_1(pr_1(T))&=(p_1(A_3),f_1(A_3)).
\end{align*}
The structures of classes $A_1$, $A_2$ and $A_3$ in the heterogeneous class $T$ can be expressed as follows:
\begin{align*}
&A_1=Core(T),\\
&A_2=Core(T)\cup pr_1(T),\\
&A_3=Core(T)\cup pr_1(T)\cup pr_1(pr_1(T)).
\end{align*}

\subsection{Multiple Inheritance}
Suppose we have the following sequence of inheritance
\[A_3\xrightarrow[]{inherits}A_1\ \ \ and\ \ \ A_3\xrightarrow[]{inherits}A_2.\]
The result of such inheritance process is 
\begin{gather*}
A_3\xrightarrow[]{inherits}A_1\ \ \ and\ \ \ A_3\xrightarrow[]{inherits}A_2=T=\\
=(pr_1(T),pr_2(T),pr_1(pr_1(T),pr_2(T))),
\end{gather*}
where
\begin{align*}
pr_1(T)&=(p_1(A_1),p_2(A_1),f_1(A_1),f_2(A_1)),\\
pr_2(T)&=(p_1(A_2),p_2(A_2),f_1(A_2)),\\
pr_1(pr_1(T),pr_2(T))&=(p_1(A_3),f_1(A_3)).
\end{align*}
The structures of classes $A_1$, $A_2$ and $A_3$ in the heterogeneous class $T$ can be expressed as follows:
\begin{align*}
&A_1=pr_1(T),\\
&A_2=pr_2(T),\\
&A_3=pr_1(pr_1(T),pr_2(T)).
\end{align*}

\subsection{Special Cases}
Let us consider example of partial and weak inheritance, using previously described classes $A_1$, $A_2$ and $A_3$ for it. Suppose we have the situation, when the class $A_2$ partially inherits the class $A_1$, for example property $p_1(A_1)$ and method $f_1(A_1)$.
\[A_2\xrightarrow[]{inherits\ (p_1, f_1)}A_1=T=(Core(T),pr_1(T),pr_2(T)),\]
where
\begin{align*}
Core(T)&=(p_1(A_1),f_1(A_1)),\\
pr_1(T)&=(p_2(A_1),f_2(A_1)),\\
pr_2(T)&=(p_1(A_2),p_2(A_2),f_1(A_2)).
\end{align*}
The structures of classes $A_1$ and $A_2$ in the heterogeneous class $T$ can be expressed as follows:
\begin{align*}
&A_1=Core(T)\cup pr_1(T),\\
&A_2=Core(T)\cup pr_2(T).
\end{align*}
Suppose we have the situation, when the class $A_2$ weakly inherits the class $A_1$, for example property $p_1(A_1)$ with measure $0.5$.
\[A_2\xrightarrow[]{inherits\ (p_1/0.5)}A_1=T=(Core(T),pr_1(T),pr_2(T)),\]
where
\begin{align*}
Core(T)&=(p_2(A_1),f_1(A_1),f_2(A_1)),\\
pr_1(T)&=(p_1(A_1)/1),\\
pr_2(T)&=(p_1(A_1)/0.5,p_1(A_2),p_2(A_2),f_1(A_2)).
\end{align*}
The structures of classes $A_1$ and $A_2$ in the heterogeneous class $T$ can be expressed as follows:
\begin{align*}
&A_1=Core(T)\cup pr_1(T),\\
&A_2=Core(T)\cup pr_2(T).
\end{align*}

\section{Conclusions}
This paper contains analysis of inheritance process and its specifics in the context of knowledge representation within OOP, frames and OODN. Such kinds of inheritance problems as problems of exceptions, redundancy, ambiguity and some approaches for their solving were considered in different perspectives. In addition, the various kinds of inheritance classifications were considered.

New types of inheritance, which allow building of inheritance hierarchy in more flexible and efficient way, were proposed. Furthermore, general classification of all known inheritance types, which includes eight different types of inheritance, was introduced. The application of approach, which allows avoiding in many cases problems with exceptions, redundancy and ambiguity within OODN and FOODN was shown, using examples. 

Proposed approach for organizing of inheritance hierarchies suggests new concepts, which can extend the OOP in many useful directions. However, despite all its benefits, it requires further research.
%
%

%
%

\end{document}